\documentclass{article}

\usepackage{cite}
\usepackage{hyperref}

\usepackage{tabularx}
\usepackage{makecell}

\newcolumntype{Y}{>{\centering\arraybackslash}X}
\usepackage{array}
\usepackage{booktabs}
\usepackage[ruled]{algorithm2e}
\usepackage{graphicx,grffile}
\makeatletter
\def\maxwidth{\textwidth}
\def\maxheight{\ifdim\Gin@nat@height>\textheight\textheight\else\Gin@nat@height\fi}
\makeatother
\setkeys{Gin}{width=\maxwidth,height=\maxheight,keepaspectratio}
\usepackage{float}
\let\origfigure=\figure
\let\endorigfigure=\endfigure
\renewenvironment{figure}{%
\origfigure[!h]
}{%
\endorigfigure
}
\providecommand{\tightlist}{%
  \setlength{\itemsep}{0pt}\setlength{\parskip}{0pt}}
\usepackage{amsmath}
\usepackage{amssymb}
\usepackage{bussproofs}
\usepackage[utf8]{inputenc}
\makeatletter

\newtheorem{definition}{Definition}
\newtheorem{example}{Example}
\newtheorem{remark}{Remark}
\makeatother

\DeclareRobustCommand{\href}[2]{#2\footnote{\url{#1}}}
\usepackage{authblk}

\begin{document}

\title{On Dedicated CDCL Strategies for PB Solvers}

\author[1]{Daniel Le Berre}
\author[,2]{Romain Wallon\thanks{Most of this paper is based on
research conducted by this author while he was working as a PhD student
at CRIL (Univ Artois \& CNRS).}}

\affil[1]{Univ. Artois, CNRS, Centre de Recherche en Informatique de Lens (CRIL), F-62300 Lens, France\linebreak\texttt{leberre@cril.fr}}
\affil[2]{LIX, Laboratoire d'Informatique de l'X, Ecole Polytechnique, X-Uber Chair\linebreak\texttt{wallon@lix.polytechnique.fr}}

\maketitle
\begin{abstract}
Current implementations of pseudo-Boolean (PB) solvers
working on native PB constraints are based on the CDCL architecture
which empowers highly efficient modern SAT solvers. In particular, such
PB solvers not only implement a (cutting-planes-based) conflict analysis
procedure, but also complementary strategies for components that are
crucial for the efficiency of CDCL, namely branching heuristics, learned
constraint deletion and restarts. However, these strategies are mostly
reused by PB solvers without considering the particular form of the PB
constraints they deal with. In this paper, we present and evaluate
different ways of adapting CDCL strategies to take the specificities of
PB constraints into account while preserving the behavior they have in
the clausal setting. We implemented these strategies in two different
solvers, namely \emph{Sat4j} (for which we consider three
configurations) and \emph{RoundingSat}. Our experiments show that these
dedicated strategies allow to improve, sometimes significantly, the
performance of these solvers, both on decision and optimization
problems.

\end{abstract}

\hypertarget{introduction}{%
\section{Introduction}\label{introduction}}

The success of so-called \emph{modern} SAT solvers has motivated the
generalization of the \emph{conflict-driven clause learning} (CDCL)
architecture \cite{grasp, chaff, minisat} to solve
\emph{pseudo-Boolean} (PB) problems~\cite{handbook}. The main
motivation behind the development of PB solvers is that classical SAT
solvers are based on the \emph{resolution} proof system, which is a
\emph{weak} proof system: instances that are hard for resolution (for
instance those requiring counting capabilities, such as \emph{pigeonhole
principle formulae} \cite{haken85}) are hard for SAT solvers. A
stronger alternative is the \emph{cutting planes} proof system
\cite{gomory58, hooker88, nordstrom15}, which allows, for instance, to
solve pigeonhole principle formulae with a linear number of derivation
steps. Generally speaking, this proof system \emph{p-simulates}
resolution: any resolution proof can be simulated by a polynomial size
cutting planes proof \cite{cook87}. In theory, PB solvers should thus
be able to find shorter unsatisfiability proofs, and thus be more
efficient than classical SAT solvers. In practice however, current PB
solvers fail to keep the promises of the theory. In particular, most
PB~solvers \cite{pbchaff, galena, pueblo, sat4j} implement a subset of
the cutting planes proof system known as \emph{generalized resolution}
\cite{hooker88}. This subset is convenient as it allows to extend the
CDCL algorithm to PB constraints. As soon as a constraint becomes
conflicting, the generalized resolution rule is applied between this
constraint and the reason for the propagation of one of its literals to
derive a new conflicting constraint. This operation is repeated until an
assertive constraint is eventually derived. However, solvers
implementing this procedure do not exploit the full power of the cutting
planes proof system \cite{jakobproofsolvers}, and are still behind
resolution-based solvers in PB competitions \cite{pb16}.

Despite the recent improvements brought by \emph{RoundingSat} \cite{rs}
with the use of the \emph{division} rule during conflict analysis,
current implementations of cutting planes still have a critical
drawback: they degenerate to resolution when given a CNF as input.
Moreover, such implementations are more complex than just replacing
resolution during conflict analysis by generalized resolution: finding
\emph{which} rules to apply and \emph{when} is not that obvious
\cite{divvssat, weakening}. In particular, PB solvers need to take care
about the specific properties of PB constraints and of the cutting
planes proof system to fit in the CDCL architecture. Additionally, CDCL
comes with many other features, without which the performance of the
solver may become very bad (see, e.g., \cite{cdclpractice}). To the
best of our knowledge, little work has been done on extending these
components for PB solvers: they are mostly reused from their definition
in classical SAT solvers, and adapted just enough to work in the solver,
without considering their effective impact in the context of PB~solving.
In this paper, we focus on such features, namely branching heuristics,
learned constraint deletion strategies and restart schemes. We
implemented different new strategies for these features, designed to
consider the characteristics of PB constraints. Our experiments show
that they allow to improve, sometimes significantly, the performance of
different PB solvers, both on decision and optimization instances.

\hypertarget{preliminaries}{%
\section{Preliminaries}\label{preliminaries}}

We consider a propositional setting defined on a finite set of
propositional variables \(\mathcal{V}\). A \emph{literal}~\(\ell\) is a
variable \(v \in \mathcal{V}\) or its negation \(\bar{v}\). Boolean
values are represented by the integers~\(1\)~(true) and \(0\)~(false),
so that \(\bar{v} = 1 - v\).

A \emph{pseudo-Boolean (PB) constraint} is an integral linear equation
or inequation over Boolean variables of the form
\(\sum_{i = 1}^{n} \alpha_i \ell_i \vartriangle \delta\), in which the
\emph{coefficients}~\(\alpha_i\) and the \emph{degree}~\(\delta\) are
integers, \(\ell_i\) are literals and
\(\vartriangle \in \{ <, \leq, =, \geq, >\}\). Such a constraint can be
\emph{normalized} in linear time into a conjunction of constraints of
the form \(\sum_{i = 1}^{n} \alpha_i \ell_i \geq \delta\) in which the
coefficients and the degree are all positive integers. In the following,
we thus assume that all PB constraints are normalized. A
\emph{cardinality constraint} is a PB constraint in which all
coefficients are equal to~\(1\) and a \emph{clause} is a cardinality
constraint of degree \(1\). This definition illustrates that
PB~constraints are a generalization of clauses, and that clausal
reasoning is thus a special case of PB~reasoning.

PB solvers have thus been designed to extend the CDCL algorithm of
classical SAT solvers. In particular, when looking for a solution, PB
solvers have to \emph{assign} variables. In the following, we use the
notation \(\ell (V@D)\) to represent that literal~\(\ell\) has been
assigned value \(V\) at decision level \(D\), and \(\ell (?@?)\) to
represent that \(\ell\) is unassigned. Assigning variables is achieved
either by \emph{making a decision} or by \emph{propagating} a truth
value for a variable. In this context, the normalized form of PB
constraints is particularly useful for detecting propagations: as for
clauses, propagations are triggered after the falsification of some
literals in the constraint. However, contrary to clauses, a PB
constraint may propagate a literal even if some other literals in this
constraint are unassigned or satisfied, as shown in the following
example.

\begin{example}
\label{ex:assignments}
The PB constraint $5a (0@3) + 5b (?@?) + c (?@?) + d (?@?) + e (0@1) + f (1@2) \geq 6$
propagates the literal $b$ under the current partial assignment.
If $b$ is assigned to $0$, giving
$5a (0@3) + 5b (0@3) + c (?@?) + d (?@?) + e (0@1) + f (1@2) \geq 6$, the constraint 
becomes conflicting.
In both cases, observe that $f$ is satisfied and $c$ and $d$ are unassigned.
\end{example}

After propagations are triggered, it may happen that a constraint
becomes conflicting. When this is the case, PB solvers perform a
conflict analysis similar to that of SAT solvers, and successively apply
the \emph{cancellation} rule between the conflicting constraint and the
reason for the propagation of some of its literals, so as to eliminate
these literals. However, doing so does not guarantee to preserve the
conflict, and several approaches based on the \emph{(partial) weakening}
rule have been introduced \cite{dixon04, rs, weakening} to provide such
a guarantee, by (locally) assuming that some literals are assigned to
\(1\). Some solvers such as \emph{Sat4j-GeneralizedResolution}
\cite{sat4j} apply this rule iteratively until the conflict is
guaranteed to be preserved, while others such as \emph{RoundingSat}
\cite{rs}, \emph{Sat4j-RoundingSat} and \emph{Sat4j-PartialRoundingSat}
\cite{weakening} apply it on \emph{all} literals that are not falsified
and not divisible by the coefficient of the literals to eliminate,
before applying the division rule.

\hypertarget{branching-heuristics}{%
\section{Branching Heuristics}\label{branching-heuristics}}

An important component in a SAT solver is its \emph{branching
heuristic}: to find efficiently a solution or an unsatisfiability proof,
the solver has to choose the \emph{right} variables on which to make
decisions. Currently, most SAT solvers rely on VSIDS \cite{chaff} or
one of its variants \cite{Biere2015}, or the more recent\linebreak LRB~\cite{Liang2016}. We focus on the former, as it is the one adopted by
the native PB solvers we considered.

The most popular variant of VSIDS is \emph{exponential VSIDS} (EVSIDS),
introduced in \emph{MiniSat} \cite{minisat}. In this heuristic, a value
\(g\) is chosen between \(1.01\) and \(1.2\) at the beginning of the
execution of the solver. When a variable is encountered during the
analysis of the \(i\)-th conflict, this variable is \emph{bumped}, i.e.,
its score is updated by adding \(g^i\) to its current score. When it
comes to selecting a variable, the solver chooses the variable with the
highest score. We remark that, as the original VSIDS, EVSIDS is designed
to favor variables appearing in recent conflicts. Moreover, modern
implementations of VSIDS not only update the score of variables
appearing in the learned clauses, but also that of variables appearing
in \emph{all} clauses used to produce them. This approach aims to favor
the selection of variables that are \emph{involved} in recent conflicts.

\hypertarget{vsids-in-pb-solvers}{%
\subsection{VSIDS in PB Solvers}\label{vsids-in-pb-solvers}}

Current PB solvers rely thus on the VSIDS heuristic (or one of its
variants) to decide which variable should be assigned next. In practice,
this heuristic may be used as is by PB solvers, even though doing so
does not allow to take into account all the information given by a PB
constraint, as observed in \cite{galena} (which, however, does not
explicitly provide a more suitable heuristic). This is why different
variants of this heuristic have been proposed. In
\cite[Section~4.5]{dixon04}, it is proposed to add, for each variable appearing in a
\emph{cardinality constraint of the original problem} (i.e., not for
\emph{learned} constraints) the degree of this constraint to the
\emph{initial} score of the corresponding variables. This approach
actually counts the occurrences of the variable in the clauses that are
represented by the cardinality constraint.

\begin{example}[{from \cite[Section 4.5]{dixon04}}]
If the cardinality constraint $a + b + c \geq 2$ is present in the original
constraint database, the score of each of its variables is increased by $2$.
Indeed, this constraint is equivalent to the conjunction of the clauses
$a + b \geq 1$, $a + c \geq 1$ and $b + c \geq 1$.
If this constraint is learned, the corresponding scores are only increased by
$1$.
\end{example}

Despite providing a more specific heuristic than the original VSIDS
heuristic when considering PB problems, this heuristic is not completely
satisfactory, as it does not fit well in modern implementations of
VSIDS, and especially of EVSIDS. First, as only the original constraints
are considered, the heuristic does not bring any improvement over the
classical implementation of the heuristic, which essentially relies on
the bumping of variables involved in recent conflicts. Second, the
particular form of general PB constraints is not taken into account by
this heuristic. The main reason for only considering cardinality
constraints in this case is that computing the number of clauses in
which a literal of a PB constraint appears is hard in general. Another
alternative, implemented in \emph{Pueblo} \cite{pueblo}, is estimating
the relative importance of a literal in a constraint, by computing the
ratio of its coefficient by the degree of the constraint. This value is
then added to the VSIDS score of the variable. On the contrary,
\emph{Sat4j} \cite{sat4j} and \emph{RoundingSat} \cite{rs} both
implement a more classical EVSIDS heuristic, by bumping each variable
encountered during conflict analysis. However, some implementation
details are worth noting for these two solvers. In particular,
\emph{Sat4j} bumps these variables \emph{each time they appear in a
reason}, while \emph{RoundingSat} bumps them only once (as in
\emph{MiniSat} \cite{minisat}), except if the variable is eliminated
during conflict analysis, in which case it is bumped \emph{twice}.

\hypertarget{towards-better-vsids-for-pb-solvers}{%
\subsection{Towards Better VSIDS for PB
Solvers}\label{towards-better-vsids-for-pb-solvers}}

As mentioned above, current implementations of the VSIDS heuristic in
SAT solvers, and in particular the EVSIDS heuristic, are designed to
favor the selection of variables that are involved in recent conflicts.
When only considering clauses, identifying such literals is
straightforward: the literals involved in a conflict are those appearing
in the clauses encountered during conflict analysis. However, this is no
longer the case when PB constraints are considered. Indeed, given a PB
constraint, the literals it contains may not play the same role in the
constraint, and thus may not have the same influence in the conflicts in
which this constraint is involved. In order to take into account this
asymmetry between the literals when computing VSIDS scores, we introduce
different ways of bumping the variables appearing in the constraints
encountered during conflict analysis. The main reason for the asymmetry
of the literals in a PB constraint is the presence of coefficients in
the constraint. To take these literals into account, we generalize the
heuristics proposed in the PB solvers \emph{pbChaff} \cite[Section
4.5]{pbchaff, dixon04} and \emph{Pueblo} \cite{pueblo} by defining the
following bumping strategies:

\begin{itemize}
\item
  The \emph{bump-degree} strategy multiplies the increment by the
  \emph{degree} of the constraint, as a naive generalization of
  \emph{pbChaff}'s approach, which only considers the degree of the
  original cardinality constraints.
\item
  The \emph{bump-coefficient} strategy multiplies the increment by the
  \emph{coefficient} of the literal being bumped, as a tentative measure
  of the importance of the corresponding variable.
\item
  The \emph{bump-ratio-coefficient-degree} strategy multiplies the
  increment by the \emph{ratio} of the coefficient of the literal by the
  degree of the constraint, as proposed in \emph{Pueblo}.
\item
  The \emph{bump-ratio-degree-coefficient} strategy multiplies the
  increment by the \emph{ratio} of the degree of the constraint by the
  coefficient of the literal, as a generalization of \emph{pbChaff}'s
  strategy taking into account the relative importance of the variable
  in the constraint.
\end{itemize}

Let us illustrate these different strategies by the following example.

\begin{example}
When bumping the variable $a$ from the constraint
$5a + 5b + c + d + e + f \geq 6$, the increment is multiplied by:
\begin{itemize}
\item $6$ in the case of \emph{bump-degree},
\item $5$ in the case of \emph{bump-coefficient},
\item $5/6$ in the case of \emph{bump-ratio-coefficient-degree} (as in \emph{Pueblo}), and
\item $6/5$ in the case of \emph{bump-ratio-degree-coefficient}
\end{itemize}
\noindent before being added to the variable's score.
\end{example}

Another key observation to take into account to detect literals that are
actually \emph{involved} in a conflict is to consider the impact of the
current assignment. Indeed, in classical SAT solvers, all variables
appearing in the clauses encountered during conflict analysis are always
\emph{assigned}, and all but one are actually \emph{falsified}. However,
in PB constraints, this is not always the case (see
Example~\ref{ex:assignments}), and falsified literals may even be
\emph{ineffective} \cite[Section 3.1]{weakening}.

\begin{definition}[Effective Literal]
Given a conflicting (resp.~assertive) PB constraint
$\chi$, a literal $\ell$ of $\chi$ is said to be \emph{effective} in
$\chi$ if it is falsified and satisfying it would not preserve the
conflict (resp. propagation).
We say that $\ell$ is \emph{ineffective} when it is not effective.
\end{definition}

\begin{remark}
To identify ineffective literals in a constraint, we use a greedy algorithm
that works as follows.
The literals of the constraint are successively (and implicitly) weakened away,
and only those for which the weakening does not preserve the conflict (resp.
propagations) are kept.
This operation, yields an (implicit) clause that is both implied by the
constraint and conflicting (resp. assertive).
Its literals are those considered as effective.
Note that this approach is similar to that used by \emph{SATIRE}
\cite{sakallah01} or \emph{Sat4j-Resolution} \cite{sat4j} to derive clauses
during conflict analysis.
\end{remark}

Even though they may be encountered during conflict analysis,
ineffective literals do not play any role in the conflict, and neither
do the corresponding variables. We thus introduce three other bumping
strategies taking into account the current assignment:

\begin{itemize}
\item
  The \emph{bump-assigned} strategy bumps only assigned variables
  appearing in the constraints encountered during conflict analysis.
\item
  The \emph{bump-falsified} strategy bumps only variables whose literals
  appear as falsified in the constraints encountered during conflict
  analysis.
\item
  The \emph{bump-effective} strategy bumps only variables whose literals
  are effective in the constraints encountered during conflict analysis.
\end{itemize}

\pagebreak

\begin{example}
When bumping the variables of the constraint
$5a (0@3) + 5b (1@3) + c (?@?) + d (?@?) + e (0@1) + f (1@2) \geq 6$,
\begin{itemize}
\item the strategy \emph{bump-assigned} bumps the variables $a$, $b$, $e$ and $f$,
\item the strategy \emph{bump-falsified} bumps the variables $a$ and $e$, and
\item the strategy \emph{bump-effective} bumps only the variable $a$.
\end{itemize}
\end{example}

\hypertarget{learned-constraint-deletion}{%
\section{Learned Constraint
Deletion}\label{learned-constraint-deletion}}

PB solvers, similarly to SAT solvers, need to regularly \emph{delete}
learned constraints during their execution. Indeed, storing these
constraints may not only increase the memory required by the solver, but
may also slow down unit propagation. In this context, the key element is
to detect \emph{which} constraints to remove. In PB solvers, this
feature is mostly inherited directly from SAT solvers. For instance,
\emph{Pueblo} \cite{pueblo} uses \emph{MiniSat}'s learned constraint
deletion, based on the activity of learned constraints (the less active
constraints are removed first), \emph{Sat4j} \cite{sat4j} uses also an
activity-based strategy but more aggressively as in
\emph{Glucose}~\cite{lbd}, while \emph{RoundingSat}~\cite{rs}
considers a custom hybrid approach, based on both the \(LBD\) and the
activity measures (the latter is used as a tie-break rule when the
former gives identical measures). In other PB solvers, such as
\emph{pbChaff} \cite{pbchaff} and \emph{Galena} \cite{galena}, the
learned constraint deletion in use (if any) is not documented. In
\cite{galena}, a perspective is however mentioned to weaken learned
constraints instead of removing them. However, note that while measures
such as those based on the activity may be reused as they are by PB
solvers (they do not take into account the representation nor the
semantics of the constraints they evaluate), for other evaluation
schemes, paying attention to the particular form of PB constraints may
be more relevant to properly evaluate the quality of the constraints.
This section focuses on two main approaches towards this direction.

\hypertarget{size-based-measures}{%
\subsection{Size-Based Measures}\label{size-based-measures}}

In classical SAT solvers, size-based measures delete the largest clauses
in the database, i.e., those containing many literals. The intuition
behind this evaluation scheme is that large clauses are weak, especially
from a propagation viewpoint: a propagation can only be triggered after
many literals have become falsified. When considering PB constraints,
this is not the case anymore. Indeed, recall that PB constraints may
propagate literals while some other literals remain unassigned, and that
the number of literals in a PB constraint does not necessarily reflect
its strength.

Another reason that motivated the use of size-based measures in SAT
solving is that large clauses are expensive to handle, which is also
true for PB constraints. In particular, in such constraints, the size
also takes into account the size of the coefficients, which is not
negligible: as coefficients may become very large during conflict
analysis, arbitrary precision encoding is required to represent these
coefficients. As we already discussed, this representation slows down
arithmetic operations, and thus the conflict analysis performed by the
solver. Different approaches have been studied to limit the growth of
the coefficient, such as those based on the division \cite{rs} or the
weakening \cite{weakening} rules. However, these approaches lead to the
inference of weaker constraints. By using a quality measure that takes
into account the size of the coefficients, we can favor the learning of
constraints with ``small'' coefficients. Towards this direction, we
introduce quality measures based on the degree of the learned
constraints, as described below:

\begin{itemize}
\item
  The \emph{degree} quality measure evaluates the quality of a learned
  constraint by the \emph{value} of its degree.
\item
  The \emph{degree-bits} quality measure evaluates the quality of a
  learned constraint by the \emph{minimum number of bits} required to
  represents its degree
\end{itemize}

In both cases, the smaller the degree, the better the constraint.
Indeed, it is well-known that the degree of a PB constraint can be used
as an upper bound of the coefficients of the constraints (because of the
\emph{saturation} rule), so that considering only the degree is enough
for the purpose of this measure.

\begin{example}
The degree-based quality measures for the constraint
$5a + 5b + c + d + e + f \geq 6$ are:
\begin{itemize}
\item $6$ in the case of \emph{degree}, and
\item $3$ in the case of \emph{degree-bits} (as the binary representation of $6$,
      i.e., $110$, needs $3$ bits).
\end{itemize}
\end{example}

\hypertarget{lbd-based-measures}{%
\subsection{LBD-Based Measures}\label{lbd-based-measures}}

Another alternative to measure the quality of learned clauses in SAT
solvers is the so-called \(LBD\) \cite{lbd}.

\begin{definition}[$LBD$]
\label{def:lbd}
Consider a \emph{clause} $\gamma$ and the current assignment of
its literals.
Let $\pi$ be a partition of these literals, such that literals are
partitioned w.r.t.~their decision levels.
The $LBD$ of $\gamma$ is the number of classes in $\pi$.
\end{definition}

The \(LBD\) of a clause is first computed when this clause is learned,
and is then updated each time the clause is used as a reason. In this
context, the notion of \(LBD\) relies on the fact that all literals in a
conflicting clause are falsified, and when the clause is used as a
reason, only one literal is not falsified (the propagated literal), but
its decision level is also that of another (falsified) literal, which
has triggered the propagation. When PB constraints are considered, this
is not the case anymore. As such, \(LBD\) is not well-defined for such
constraints. To consider it as a quality measure for learned PB
constraints, we thus need to take into account the literals that are
unassigned in these constraints. To do so, we introduce five different
definitions of this measure. First, we consider a sort of default
definition of \(LBD\) for PB constraints, which only takes into account
assigned literals. This definition of \(LBD\) was used for instance in
the first version of \emph{RoundingSat} \cite{rs}.

\begin{definition}[$LBD_a$]
Consider a PB constraint $\chi$ and the current assignment of
its \emph{assigned} literals.
Let $\pi$ be a partition of these literals, such that literals are
partitioned w.r.t.~their decision levels.
The $LBD_a$ of $\chi$ is the number of classes in $\pi$
(``$a$'' stands for ``assigned'').
\end{definition}

Unassigned literals may be considered as if they were assigned to a
``dummy'' decision level. This decision level may be the same for all
literals, or not.

\begin{definition}[$LBD_s$]
Consider a PB constraint $\chi$ and the current assignment of
its \emph{assigned} literals.
Let $\pi$ be a partition of these literals, such that literals are
partitioned w.r.t.~their decision levels.
Let $n$ be the number of classes in $\pi$.
The $LBD_s$ of $\chi$  is $n$ if all literals in $\chi$ are assigned, and
$n + 1$ otherwise (``$s$'' stands for ``same'').
\end{definition}

\begin{definition}[$LBD_d$]
Consider a PB constraint $\chi$ and the current assignment of
its \emph{assigned} literals.
Let $\pi$ be a partition of these literals, such that literals are
partitioned w.r.t.~their decision levels.
Let $n$ be the number of classes in $\pi$.
The $LBD_d$ of $\chi$  is $n + u$, where $u$ is the number of unassigned
literals in $\chi$ (``$d$'' stands for ``different'').
\end{definition}

Another possible extension of \(LBD\) is to only consider falsified
literals, as in the current version of \emph{RoundingSat}:

\begin{definition}[$LBD_f$]
Consider a PB constraint $\chi$ and the current assignment of
its \emph{falsified} literals.
Let $\pi$ be a partition of these literals, such that literals are
partitioned w.r.t.~their decision levels.
The $LBD_f$ of $\chi$ is the number of classes in $\pi$
(``$f$'' stands for ``falsified'').
\end{definition}

The definition above is based on the observation that, when a clause is
learned, all literals in this clause are falsified. However, it may
happen that falsified literals in a PB constraint are actually
\emph{ineffective} (while this is never the case in a clause). As these
literals are not involved in the conflict, we should not consider them
either. We thus define another extension of \(LBD\) that only considers
effective literals:

\begin{definition}[$LBD_e$]
Consider a PB constraint $\chi$ and the current assignment of
its \emph{effective} literals.
Let $\pi$ be a partition of these literals, such that literals are
partitioned w.r.t.~their decision levels.
The $LBD_e$ of $\chi$ is the number of classes in $\pi$
(``$e$'' stands for ``effective'').
\end{definition}

\clearpage

\begin{example}
The $LBD$-based quality measures for the constraint $\chi$ given by
$5a (0@3) + 5b (1@3) + c (?@?) + d (?@?) + e (0@1) + f (1@2) \geq 6$ are:

\begin{itemize}
\item $LBD_a(\chi) = |\{ \{a, b\}, \{e\}, \{f\} \}| = 3$
\item $LBD_s(\chi) = |\{ \{a, b\}, \{c, d\}, \{e\}, \{f\} \}| = 4$
\item $LBD_d(\chi) = |\{ \{a, b\}, \{c\}, \{d\}, \{e\}, \{f\} \}| = 5$
\item $LBD_f(\chi) = |\{ \{a\}, \{e\} \}| = 2$
\item $LBD_e(\chi) = |\{ \{a\} \}| = 1$
\end{itemize}
\end{example}

We remark that the definitions of \(LBD\) introduced in this section are
\emph{extensions} of the original definition of \(LBD\) (as given by
Definition \ref{def:lbd}), in the sense that they all coincide when
learning clauses.

\hypertarget{deleting-pb-constraints}{%
\subsection{Deleting PB Constraints}\label{deleting-pb-constraints}}

Taking advantage of the measures described above, we define the
following deletion strategies, which are applied each time the learned
clause database is reduced:

\begin{itemize}
\tightlist
\item
  \emph{delete-degree}, which deletes the constraints with the highest
  degree,
\item
  \emph{delete-degree-bits}, which deletes the constraints with the
  largest degree,
\item
  \emph{delete-lbd-a}, which deletes the constraints with the highest
  \(LBD_a\),
\item
  \emph{delete-lbd-s}, which deletes the constraints with the highest
  \(LBD_s\),
\item
  \emph{delete-lbd-d}, which deletes the constraints with the highest
  \(LBD_d\),
\item
  \emph{delete-lbd-f}, which deletes the constraints with the highest
  \(LBD_f\), and
\item
  \emph{delete-lbd-e}, which deletes the constraints with the highest
  \(LBD_e\).
\end{itemize}

\hypertarget{restarts}{%
\section{Restarts}\label{restarts}}

Restarts are a very powerful feature of CDCL SAT solvers
\cite{Gomes1998}. Even though this feature is not completely
understood, it seems required to exploit more power of the resolution
proof
system~\cite{darwichepower, jakobproofsolverspractical, AtseriasRestarts}.
Restarting is mainly forgetting all decisions made by the solver, and go
back to the root decision level. The main advantage of doing so is that
wrong decisions made at the very beginning of the search can be
cancelled to avoid being stuck in a subpart of the search space. To this
end, many restart schemes have been proposed \cite{arminrestarts},
either static such as those based on the Luby series
\cite{luby, lubyrestarts} or dynamic, as in \emph{PicoSAT}
\cite{picosatrestarts} or \emph{Glucose} \cite{glucoserestarts}. In
this section, we focus on the latter, considering restart strategies
based on the quality of learned constraints. Such restarts are not
exploited in current PB solvers. In solvers such as \emph{pbChaff}
\cite{pbchaff} or \emph{Galena} \cite{galena}, it is not clear whether
restarts are implemented or not, as they do not mention this feature. As
\emph{Pueblo} \cite{pueblo} is heavily based on
\emph{MiniSat}~\cite{minisat}, it is most likely to inherit its restart
policy, even though no mention of this feature is made in~\cite{pueblo}
either. Regarding more recent solver, \emph{Sat4j} \cite{sat4j}
implements \emph{PicoSAT}'s static and aggressive restart scheme
\cite{picosat} and \emph{RoundingSat} \cite{rs} uses a Luby-based
restart policy \cite{luby, lubyrestarts}. Note that a common point to
these two strategies is that they do not take into account the
constraints that are being considered, as they are both static policies.
They may thus be reused without any modification since they are
independent from the type of the constraints being considered. In this
section, we propose instead to follow \emph{Glucose}'s restart policy
\cite{glucoserestarts}. In this solver, the decision of whether a
restart should be performed depends on the quality of the constraints
that are currently being learned: when this quality decreases, the
solver is most likely exploring the wrong search space. As of
\emph{Glucose}, the quality of learned clauses is measured with their
\(LBD\) (see Definition~\ref{def:lbd}). To measure the decrease in the
quality of learned clauses, the average \(LBD\) is computed over the
most recent clauses (in practice, the last 100~clauses). Whenever this
average is greater than 70\% of the average \(LBD\) computed over all
learned clauses, a restart should be performed. \emph{Glucose} also
implements a wide variety of tricks to improve its restart policy (such
as restart blocking) that are beyond the scope of this paper.

We thus define 7 restarts strategies, that exploit the quality measures
defined in Section~\ref{learned-constraint-deletion}, namely
\emph{restart-degree}, \emph{restart-degree-bits},\emph{restart-lbd-a},
\emph{restart-lbd-s}, \emph{restart-lbd-d}, \emph{restart-lbd-f} and
\emph{restart-lbd-e}.

\hypertarget{experimental-results}{%
\section{Experimental Results}\label{experimental-results}}

This section presents an empirical evaluation of the different
strategies presented in this paper implemented in two PB solvers, namely
\emph{Sat4j} \cite{sat4j} and \emph{RoundingSat}~\cite{rs}. All
experiments have been executed on a cluster of computers equipped with
quadcore bi-processors Intel XEON X5550 (2.66 GHz, 8 MB cache). The time
limit was set to 1200 seconds and the memory limit to 32~GB. For space
reasons, this section does not report the results of all individual
strategies presented in this paper, but focuses on the performance of
those providing the best improvements to the considered solvers. The
interested reader may still have a look to the publicly available
detailed results of our experiments \cite{artifact}.

\hypertarget{solver-configurations}{%
\subsection{Solver Configurations}\label{solver-configurations}}

Let us first describe our implementation of the different strategies in
\linebreak \emph{Sat4j}~\cite{sat4j}, which are available in
\href{https://gitlab.ow2.org/sat4j/sat4j/tree/cdcl-strategies}{its
repository}. For this solver, we considered three main configurations,
namely \emph{Sat4j-GeneralizedResolution}, \emph{Sat4j-RoundingSat}
and \emph{Sat4j-PartialRoundingSat} \cite{weakening}. 

\pagebreak

For these three configurations, the default strategies are given below:

\begin{itemize}
\tightlist
\item
  the branching heuristic bumps all variables appearing in each
  constraint encountered during conflict analysis each time they are
  encountered,
\item
  learned constraints are stored in a \emph{mono-tiered} database, and
  are regularly deleted using \emph{MiniSat}'s learned constraint
  deletion strategy \cite{minisat}, based on the \emph{activity} of
  learned constraints (i.e., the constraints to remove are those that
  are less involved in recent conflicts), and
\item
  the restart policy is that of \emph{PicoSAT} \cite{picosat}.
\end{itemize}

Based on our experiments, the \texttt{best} combination of strategies
for \emph{Sat4j-GeneralizedResolution} is \emph{bump-effective},
\emph{delete-lbd-s} and \emph{restart-degree}, while the \texttt{best}
combination for both \emph{Sat4j-RoundingSat} and
\emph{Sat4j-PartialRoundingSat} is \emph{bump-assigned},
\emph{delete-degree-bits} and the static restart policy of \linebreak
\emph{PicoSAT}~\cite{picosat}.

\bigskip

For \emph{RoundingSat} \cite{rs}, our implementation is available in a
\href{https://gitlab.com/pb-cdcl-strategies/roundingsat/-/tree/cdcl-strategies}{dedicated
repository}. We refactored this solver starting from commit
\texttt{a17b7d0e} (denoted \texttt{master} in the following) to support
the use of the different strategies presented in this paper. The
\texttt{default} configuration of this solver corresponds to the
refactored version of \emph{RoundingSat} set up with the default
strategies originally used by this solver, i.e.:

\begin{itemize}
\tightlist
\item
  the branching heuristic bumps all variables appearing in each
  constraint encountered during conflict analysis once, and twice when
  eliminated,
\item
  learned constraints are stored in a \emph{mono-tiered} database, and
  regularly deleted using the \(LBD_f\) of the constraints and their
  activity as a tie-break, and
\item
  the restart policy uses the Luby series (with factor \(100\))
  \cite{lubyrestarts}.
\end{itemize}

The \texttt{best} combination of strategies for this solver, according
to our experiments, is \emph{bump-assigned} (with a bumping on the
variables each time they are encountered), \emph{delete-lbd-e} and
\emph{restart-lbd-e}.

\hypertarget{decision-problems}{%
\subsection{Decision Problems}\label{decision-problems}}

\begin{figure}
\centering
\includegraphics{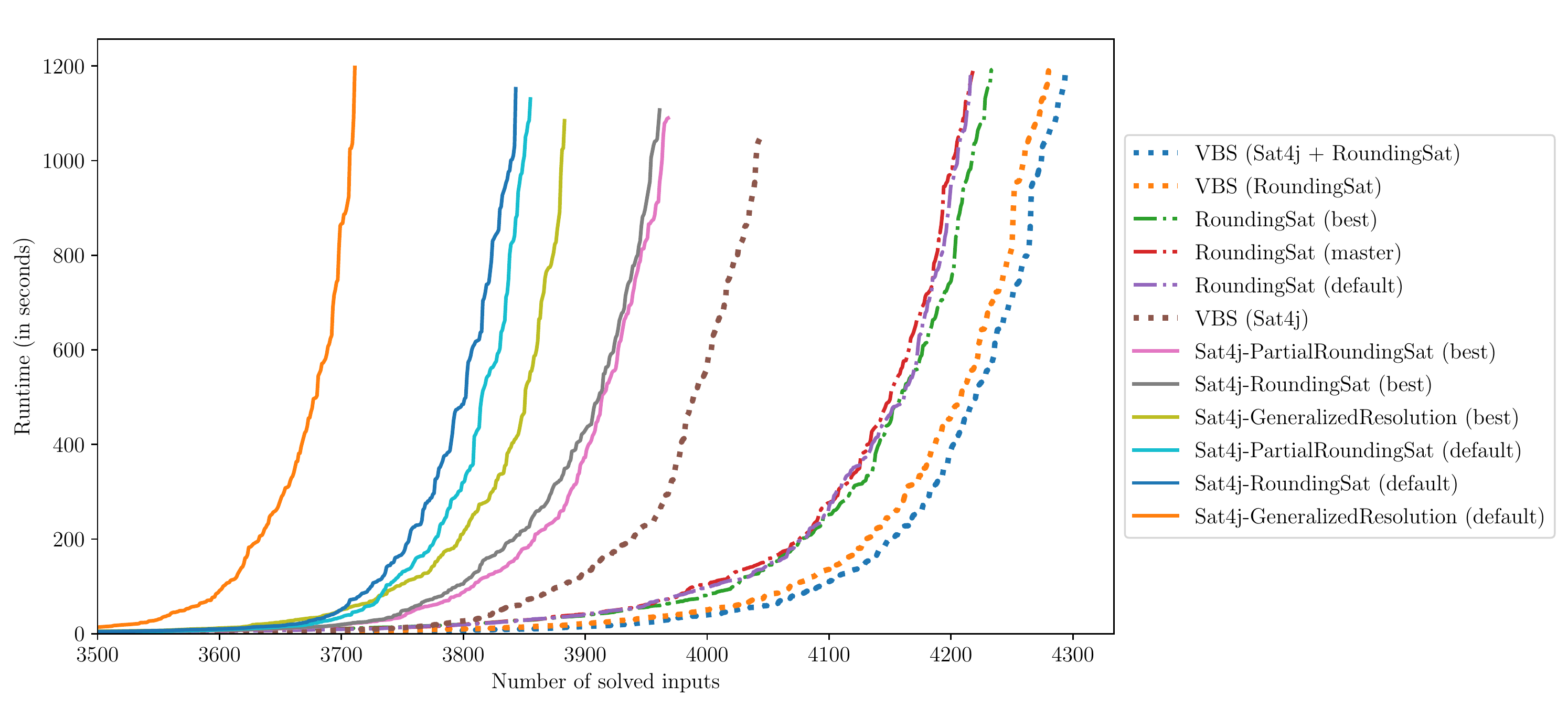}
\caption{Cactus plots of different configurations of \emph{Sat4j} and
\emph{RoundingSat} on decision problems. For more readability, the first
(easy) \(3500\) instances are cut out. \label{fig:decision}}
\end{figure}

We first consider the performance of the different solvers on decision
problems. To this end, we ran the different solvers on the whole set of
decision benchmarks containing ``small'' integers used in the PB
competitions since the first edition~\cite{pb05}, for a total of 5582
instances. Figure~\ref{fig:decision} gives the results of the different
solvers on these inputs, with their \texttt{default} and \texttt{best}
configurations.

The cactus plot shows that the different configurations of \emph{Sat4j}
are significantly improved by the use of our dedicated strategies. Quite
interestingly, we can also observe that
\emph{Sat4j-GeneralizedResolution} with the best combination of the
strategies beats both implementations of \emph{RoundingSat} in
\emph{Sat4j} with their default strategies. In the case of
\emph{RoundingSat}, we can also note a small improvement over its
default configuration, but this improvement is not as significant as in
\emph{Sat4j}. Let us remark that combining the best strategies is not
enough to get the best of all the strategies we investigated. In
particular, for each feature we considered, the \emph{Virtual Best
Solver} (\emph{VBS}) of the different strategies, i.e., the one obtained
by selecting the best performing strategies on each individual input,
has far better performance than each individual strategy, and this
applies to all configurations of \emph{Sat4j} and \emph{RoundingSat}
This suggests that no strategy is better than the other on all
benchmarks, and that they are actually complementary.

\hypertarget{optimization-problems}{%
\subsection{Optimization Problems}\label{optimization-problems}}

\begin{figure}
\centering
\includegraphics{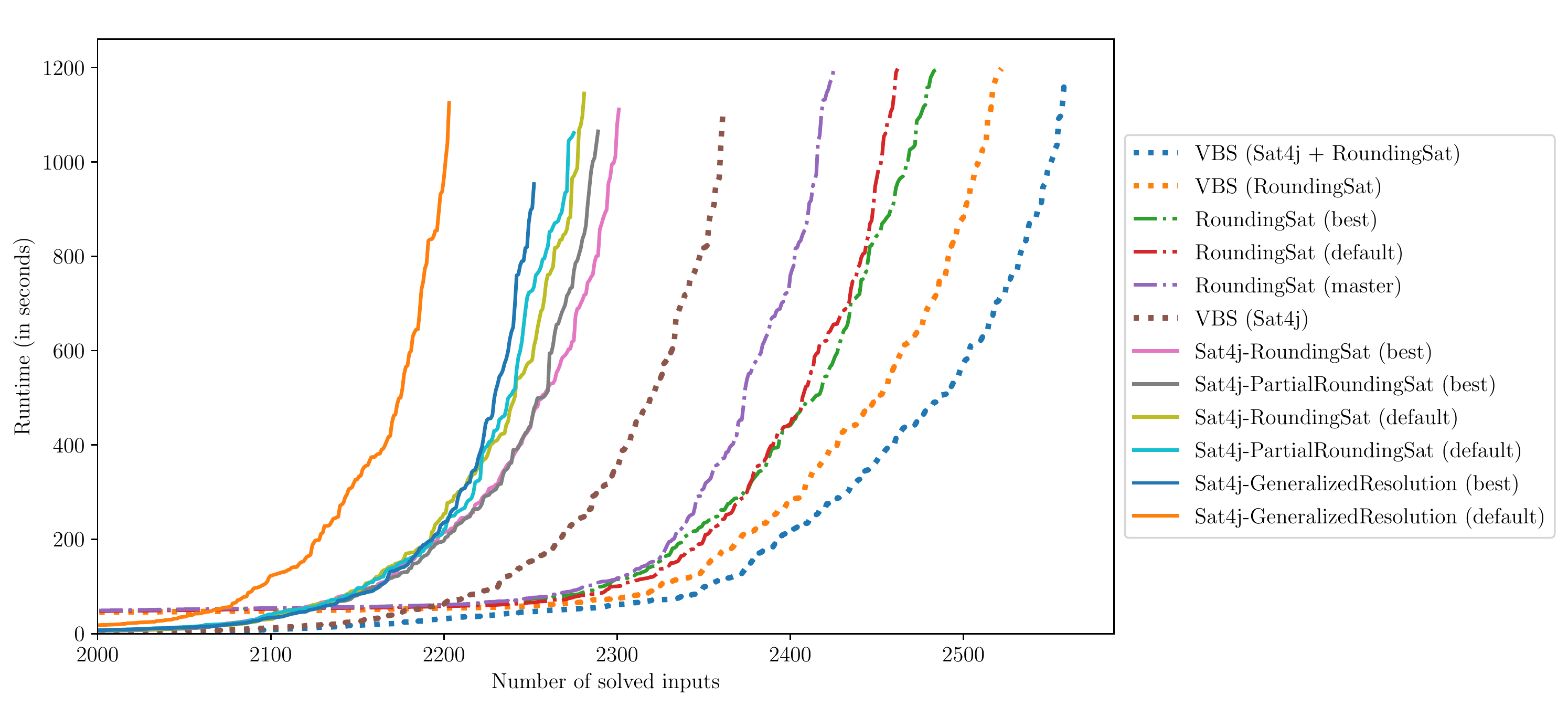}
\caption{Cactus plots of different configurations of \emph{Sat4j} and
\emph{RoundingSat} on optimization problems. For more readability, the
first (easy) \(2000\) instances are cut out. \label{fig:cactus-decision}
\label{fig:optim}}
\end{figure}

Let us now consider the performance of the different solvers on
optimization problems, by using as input the whole set of optimization
benchmarks containing ``small'' integers used in the PB competitions
since the first edition~\cite{pb05}, for a total of 4374 instances.
Considering the huge amount of computation time needed to perform our
exhaustive experiments on decision problems (more than \(8\) years of
CPU time), we focused for these experiments on the \texttt{best}
configurations of the different solvers we identified on decision
problems (which still took about \(9\) months of CPU computation time).
Figure~\ref{fig:optim} shows the results we obtained for these
configurations.

Similarly to decision problems, we can observe on the cactus plots that
all solvers are improved by the dedicated strategies on optimization
problems, with a particularly significant improvement to
\emph{Sat4j-GeneralizedResolution}.

\hypertarget{discussion}{%
\subsection{Discussion}\label{discussion}}

Let us now make a more detailed analysis of our experimental results.

Not so surprisingly, the strategy that has the most important impact,
especially in \emph{Sat4j}, is the bumping strategy, i.e., the branching
heuristic. On the one hand, our experiments showed that the strategies
\emph{bump-degree} and \emph{bump-ratio-degree-coefficient} have really
poor performance in all considered solvers (including
\emph{RoundingSat}). As described in \cite[Section~4.5]{dixon04}, these
strategies are designed to estimate the number of clauses that are
represented by the PB constraint whose literals are being bumped.
However, when a conflict occurs, not all these clauses are actually
involved in the conflict, and thus some variables get ``more bumped''
than they should be.

On the other hand, assignment-based bumping strategies are, among all
individual strategies, those having the biggest impact on the
performance of \emph{Sat4j}. For instance, we observed that
\emph{Sat4j-GeneralizedResolution} solves the (optimization) instances
of the \texttt{factor} family much faster thanks to the
\emph{bump-effective} strategy (changing the learned constraint deletion
or restart strategies makes almost no difference on this family). We
made further investigations to understand why there was such an
improvement, and it appears that the production of \emph{irrelevant
literals} (i.e., literals that occur in a PB constraint, but never
affect its truth value, whatever their assignment) penalize the solver
on this particular family. It is known that such literals may impact the
size of the proof built by PB solvers \cite{LeBerreMMW2020}. Our
experiments here also show that they may \emph{pollute} the solver's
heuristic, as \emph{bump-effective} never bumps irrelevant literals
(they are always ineffective). This also proposes another way to deal
with such literals.

The big impact of the bumping strategies in \emph{Sat4j} may also
explain why the gain in \emph{RoundingSat} is not so significant.
Indeed, the aggressive weakening performed by \emph{RoundingSat} tends,
in a sense, to already identify the literals that are already involved
in the conflict. This is particularly visible if we look at the behavior
the different bumping strategies in \emph{RoundingSat}: there is almost
no difference between them. This suggests that the gain in this solver
comes mostly from the learned constraint deletion strategy or the
restart policy, which improve the default strategies without being
significantly better.

In particular, we observed that, in \emph{Sat4j}, performing no deletion
at all is actually better than the (default) activity-based deletion
strategy. This may be explained by the fact that PB solvers are often
slower in practice than SAT solvers, especially because the operations
they need to perform, such as detecting propagations and applying the
cancellation rule, are more complex than their counterpart in SAT
solvers. This means that the number of conflicts per second in a PB
solver is lower than that in a SAT solver, and so is the number of
learned constraints. As a consequence, PB solvers do not need to clean
their learned constraint database as regularly as a SAT solver.

Regarding the restart policies, there is no big difference between the
strategies, except for \texttt{degree-bits}, which does not have good
performance compared to the others, and especially to \texttt{degree}.
This may be explained by the fact that degrees with the same number of
bits may take very different values. These are taken into account by the
latter while the former does not distinguish them. Nevertheless, there
is clearly room for improvement as the VBS performs much better than the
individual strategies.

It is also important to note that the different strategies we considered
in this paper are often tightly linked in the solver, and may thus
interact with each other. This is particularly true for the learned
constraint deletion and restart policies, since they use the same
quality measures. While using them independently does not necessarily
have a big impact on the solver (this is particularly true for the
learned constraint deletion strategy), combining them often allows to
get better performance. For instance, in \emph{RoundingSat}, while the
best (individual) strategies are \emph{PicoSAT}'s restart policy and the
deletion based on the \(LBD_s\), the best gain is actually obtained by
using the \(LBD_e\) quality measure both for learned constraint deletion
and restarts.

Another consequence of the tight link between the different strategies
and the solver itself is that implementation details may have unintended
side effects on the performance of the solver. For instance, to
implement the new strategies in \emph{RoundingSat}, we had to adapt the
code and change some data structures in the branching heuristic (by
replacing an \emph{ordered set} with an \emph{(unordered) hash map}),
resulting in the same literals being bumped, but \emph{in a different
order}. As the insertion/update order of the variables is used as a
tie-break by EVSIDS, the order in which the literals are selected varies
between the \texttt{master} and \texttt{default} configuration of the
solver, which increases the difficulty to interpret the results of
\emph{RoundingSat}, especially on optimization problems.

To conclude this analysis, let us summarize the main outcomes of our
experiments. The biggest impact on the solver is obtained by carefully
adapting the bumping strategy: while considering coefficients in this
case worsens all tested solvers, considering the current partial
assignment may drastically improve them. Regarding constraint deletion,
using the activity based measure (which is the default in \emph{Sat4j},
and only a tie-break in \emph{RoundingSat}) has really poor performance.
The other strategies have a lesser impact on the solver, and seem more
closely dependent on the proof system of the solver to bring
improvement. However, if one needs to set up strategies that work well
for all different proof systems, it would be \texttt{bump-assigned} for
the bumping strategies, \texttt{delete-lbd-s} for the deletion strategy
and either \texttt{degree}-based or \emph{PicoSAT}'s restart policy
(depending of whether big degrees are expected to be produced or not,
respectively).

\hypertarget{conclusion}{%
\section{Conclusion}\label{conclusion}}

In this paper, we introduced different branching heuristics, learned
constraint deletion and restart strategies dedicated to native PB
solving. These strategies are generalizations of those classically
implemented in SAT solvers, and are designed to take into account the
properties of PB constraints to better fit in the CDCL architecture. Our
experiments revealed that one of the key aspects of PB constraints to
take into account is the current assignment of their literals. This is
particularly true for the EVSIDS-based heuristics, but also for the
learned constraint deletion strategies and the restart policies through
the use of new \(LBD\)-based measures. When combined, these strategies
allow to improve the PB solvers \emph{RoundingSat} and \emph{Sat4j},
with a particularly significant improvement for the latter, both on
decision and optimization problems.

Nevertheless, none of these strategies performs better than the others
on all benchmarks: their VBS clearly beats each individual strategy,
even when considering their combination. Yet, the strategies introduced
in this paper show that better adapting SAT strategies may improve the
performance of PB solvers. A perspective for future research is to find
better ways to adapt such strategies, and to define new strategies that
are \emph{specifically} designed for PB solving or PB optimization
(rather than \emph{adapting} existing strategies). Another avenue to
explore is to find how to properly combine these strategies to get their
best, while taking into account the interactions between these different
strategies. In particular, it is not clear that combining all single
best strategies provides the best combination. A possible approach to
identify such a combination is to use dynamic algorithm configuration to
select the most appropriate strategies according to the state of the
solver \cite{hutterdac}.

\paragraph*{Acknowledgements.}
The authors are grateful to the anonymous reviewers for their numerous
comments, that greatly helped to improve the paper. Part of this work
was supported by the French Ministry for Higher Education and Research
and the Hauts-de-France Regional Council through the ``Contrat de Plan
État Région (CPER) DATA''. This publication was supported by the Chair
``Integrated Urban Mobility'', backed by L'X -- École Polytechnique and
La Fondation de l'École Polytechnique and sponsored by Uber. The
Partners of the Chair shall not under any circumstances accept any
liability for the content of this publication, for which the author
shall be solely liable.

\bibliography{bibliography.bib}

\end{document}